\documentclass[letterpaper, 10 pt, conference, final]{ieeeconf}  

\IEEEoverridecommandlockouts                              

\title{\LARGE \bf
Learning-based GNSS Uncertainty Quantification \\ using Continuous-Time Factor Graph Optimization
}

\author{Haoming Zhang,~\IEEEmembership{Member,~IEEE}
\thanks{Haoming Zhang was with the Institute of Automatic Control, RWTH Aachen University (Corresponding: haoming.zhang@rwth-aachen.de).}%
}

\usepackage{siunitx}
\usepackage{graphicx}
\usepackage{multirow}
\usepackage{comment}
\usepackage{bm}
\usepackage{hhline}
\usepackage[font=small,labelfont=bf]{caption}
\usepackage[noadjust]{cite}

\begin{document}
\maketitle
\thispagestyle{empty}
\pagestyle{empty}

\begin{abstract}
This short paper presents research findings on two learning-based methods for quantifying measurement uncertainties in global navigation satellite systems (GNSS). We investigate two learning strategies: offline learning for outlier prediction and online learning for noise distribution approximation, specifically applied to GNSS pseudorange observations. To develop and evaluate these learning methods, we introduce a novel multisensor state estimator that accurately and robustly estimates trajectory from multiple sensor inputs, critical for deriving GNSS measurement residuals used to train the uncertainty models. We validate the proposed learning-based models using real-world sensor data collected in diverse urban environments. 
Experimental results demonstrate that both models effectively handle GNSS outliers and improve state estimation performance. Furthermore, we provide insightful discussions to motivate future research toward developing a federated framework for robust vehicle localization in challenging environments.

\end{abstract}

\section{INTRODUCTION}
Robot navigation in large-scale outdoor environments typically relies on global navigation satellite systems (GNSS) for globally consistent state estimation. However, GNSS observations are often distorted in challenging environments, such as urban canyons, due to multipath effects and non-line-of-sight (NLOS) receptions. In such cases, GNSS measurement outliers exhibit time-varying noise dynamics characterized by heavy-tailed, skewed, and multimodal distributions that are difficult to model in advance \cite{hsu_analysis_NLOS}. Consequently, state estimators based on least-squares techniques may degrade or even diverge when relying on inconsistent measurement models with overconfident, hand-crafted noise parameters. As a result, developing a vehicle localization approach that ensures robust long-term operation across diverse environments remains a significant challenge. 

State-of-the-art methods generally tackle this challenge by leveraging four concepts: 
\subsubsection{Multisensor Fusion}\label{sec: literature_fusion}
Many approaches integrate additional sensor modalities, such as near-field ranging signals \cite{gnss_uwb}, lidar \cite{liosam}, cameras \cite{gvins, wen_vis_gnss}, or multiple sensors \cite{multisensor1}, alongside GNSS for state estimation, primarily in SLAM applications. However, these methods require precise data synchronization and careful graph construction to effectively fuse heterogeneous sensor measurements. To address this challenge and enhance estimation performance, recent approaches introduce multi-graph structures for flexible and compact sensor fusion \cite{multisensor_construction, superodom}. However, these structures add system complexity and require meticulous engineering. Other methods \cite{VI_GPS_project, multisensor_propagation} leverage high-frequency IMU measurements to align asynchronous global pose data (e.g., GNSS) with primary sensing timestamps using pre-integration. Yet, reliance on noisy IMU data still introduces uncertainty.

\subsubsection{Outlier Rejection}\label{sec:literature_outlier}
The robustness of multi-sensor fusion remains uncertain when GNSS data is severely corrupted by inconsistent noise models in challenging environments. A common approach to mitigating this issue is rejecting faulty GNSS measurements (outliers) without prior knowledge of their statistical characteristics. Methods in this category typically employ consensus checking (e.g., RANSAC-like algorithms) 
\cite{ransac_gnss_Zhang} and integrity monitoring techniques \cite{IM_GNSS_review}. However, like RANSAC, integrity monitoring assumes sufficient healthy measurements to maintain consistency. When outliers dominate or too few measurements are available, the estimated protection level becomes unreliable.

\begin{figure}[!t]
    \centering
    \includegraphics[width=0.45\textwidth]{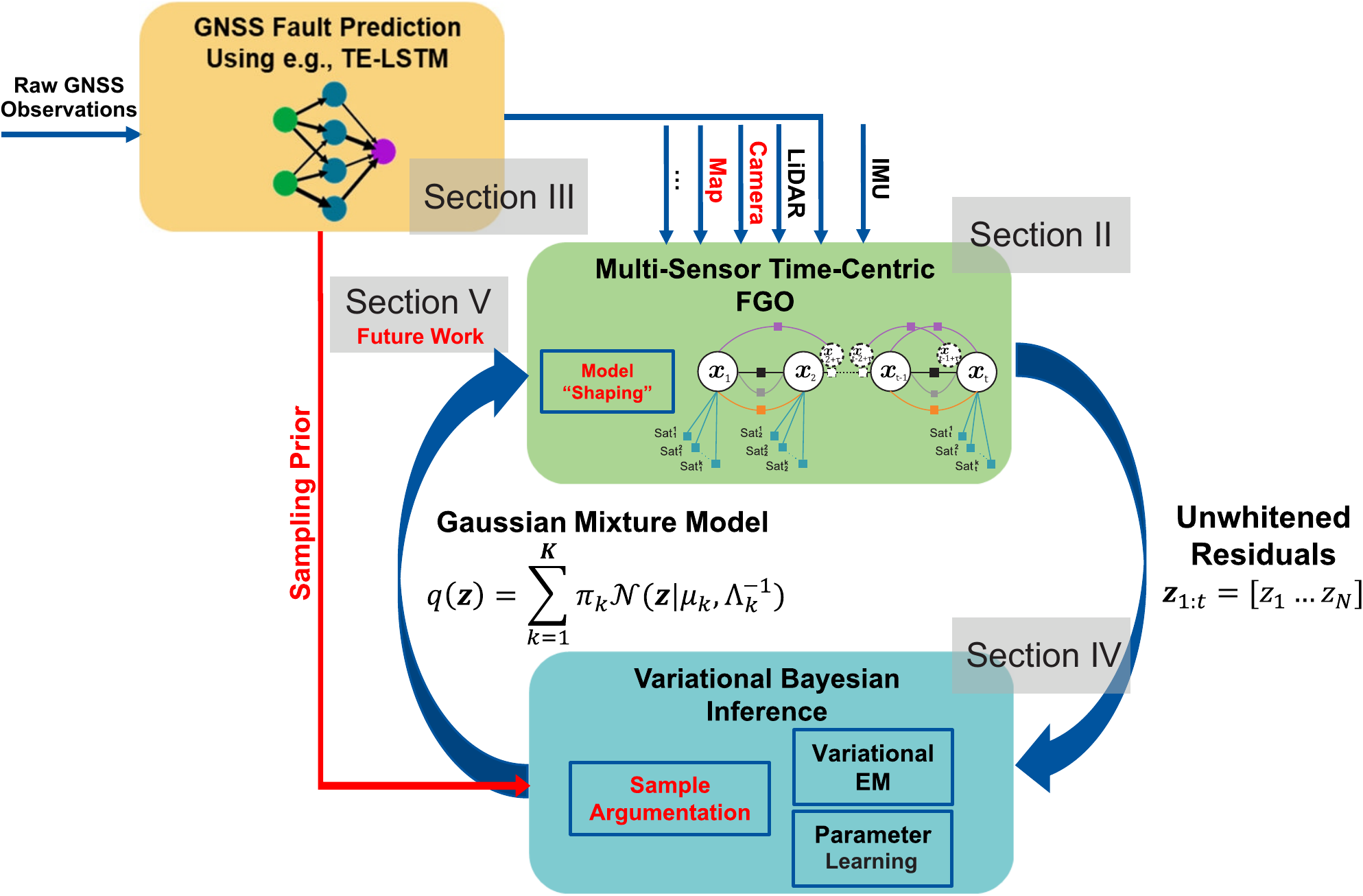}
    \caption{Schema of all proposed methods embedded in an embodied system, where the red contents indicate future work.}
    \label{fig: schema}
\end{figure}

\subsubsection{Robust Error Modeling}\label{sec: literature_error_modeling}
Robust error modeling offers another effective approach to outlier rejection by shaping cost functions using methods such as M-estimators \cite{medina_on_robust_statistics} or the Graduated Non-Convexity method \cite{fgo_gnc} within least-square estimators. Additionally, switchable constraints \cite{SwitchConvGNSS} and covariance matrix tuning \cite{DCE} are widely used in GNSS-based robot localization. While M-estimation is well-established both theoretically and practically, its effectiveness relies on selecting a well-parameterized kernel function. Moreover, these methods cannot explicitly exclude faulty GNSS measurements, leaving stability and robustness unguaranteed. In many cases, limited knowledge of measurement noise and insufficiently sampled residual data hinder accurate noise modeling for specific applications.

\subsubsection{Learning-based Methods}\label{sec: literature_learning}
Recently, learning-based methods have gained attention for handling faulty GNSS observations. Pre-trained models enable online GNSS fault detection. While classic machine-learning methods achieve high accuracy \cite{classical_learning_1, classical_learning_2}, deep learning (DL) better handles time-correlated, nonlinear GNSS data by extracting semantic features and improving generalization \cite{learning_survey, lstm_base}. Additionally, Gaussian mixture models, using Variational Bayes inference in an online learning paradigm, effectively capture noise characteristics \cite{phd_Pfeifer, Wenda_VBIGMM}.

However, while previous studies highlight the potential of learning-based methods for handling GNSS outliers, they have not thoroughly evaluated their real-world feasibility, limitations, or generalization across diverse datasets and environments. Moreover, key aspects such as spatio-temporal characteristics in GNSS signal propagation and sample augmentation remain unexplored. 

The research introduced in this paper aims to address these gaps by proposing novel methods and systematically evaluating the feasibility and limitations of learning-based approaches in real-world GNSS-aided robot navigation applications. It also explores the influence of spatio-temporal characteristics in GNSS signal propagation and the impact of data distribution shifts on model performance and generalization.
Fig.\,\ref{fig: schema} presents a systematic overview of the proposed methods. Our contributions are summarized as follows: a) We propose a generalized GNSS/multisensor state estimator; b) We propose a novel neural network architecture to detect GNSS NLOS signals and predict pseudorange errors; c) We propose an online algorithm leveraging variational Bayesian inference to approximate the noise distribution of GNSS pseudorange.

\section{Continuous-Time Factor Graph Optimization}\label{sec: fgo}
Building on our previous study in \cite{cont_fgo_zhang}, we propose a generalized factor graph optimization (FGO) by translating classic FGO for multisensor fusion into an approach using continuous-time trajectory representation where the graph associated with all to-be-estimated state variables is constructed deterministically based on a priori chosen timestamps \cite{gnssfgo}. It thus presents a time-centric factor graph construction that is independent of any particular reference sensor (e.g., GNSS). To achieve this, we represent the vehicle trajectory in continuous time using a Gaussian process (GP). 
The algorithm feeds new observations from each sensor independently into the factor graph without measurement-to-state synchronization. If a measurement cannot be temporally aligned with any state variable, we query a GP-interpolated state corresponding to the measurement used for the error evaluation.

We employed datasets from measurement campaigns in Aachen, Düsseldorf, and Cologne and presented comprehensive discussions on sensor observations, smoother types, and hyperparameter tuning. 
Our results show that the proposed approach enables robust trajectory estimation in dense urban areas where a classic multi-sensor fusion method fails due to sensor degradation. We refer the readers to \cite{gnssfgo} for detailed results. 

\textbf{Take-Away Message:}
A time-centric FGO framework does not depend on any specific sensor, thereby avoiding information loss and mitigating the risk of systematic failures, making it a more robust and flexible approach for state estimation. 

\section{Offline Learning for Outlier Prediction}\label{sec: nlos}
To address the limitations of current learning-based approaches for faulty GNSS prediction (see Section \ref{sec: literature_learning}), we take into account the noise characteristics of GNSS observations in urban environments and propose a transformer-enhanced Long short-term Memory (TE-LSTM) network in \cite{te_lstm}. 

\begin{figure*}[!t]
    \centering
    \includegraphics[width=1\textwidth]{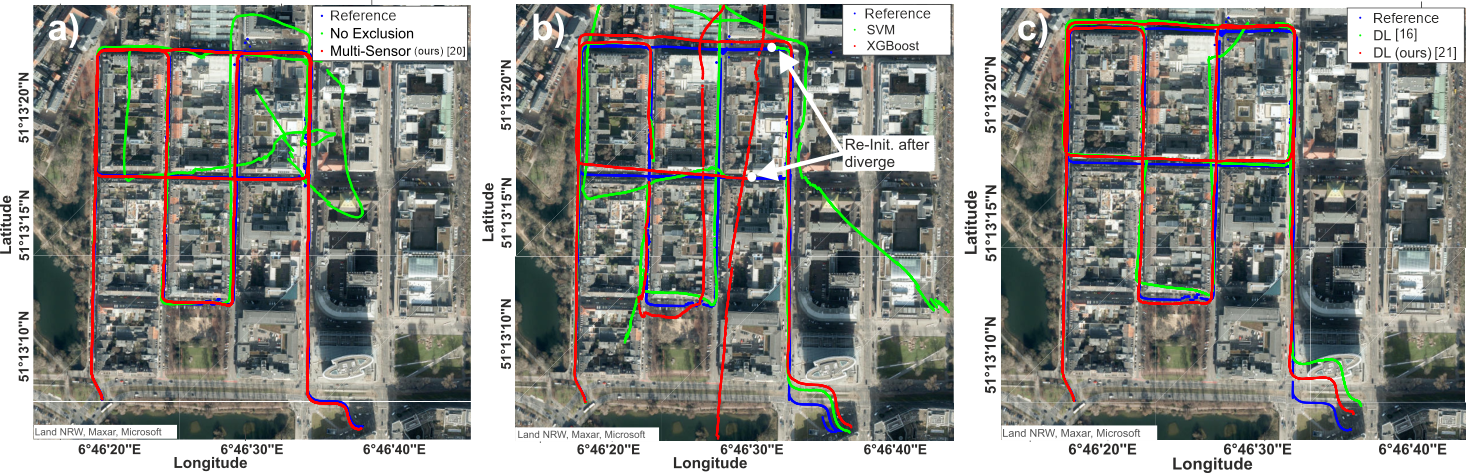}
    \caption{Vehicle Localization in Urban Area in Düsseldorf with NLOS Exclusion. The GNSS reference solution is shown in blue. a) presents the estimated trajectory without NLOS exclusion and the solution by fusing GNSS observations and lidar odometry in a tight coupling (Section \ref{sec: fgo}). b) and c) illustrate the estimated trajectories with NLOS exclusion using different learning models.}
    \label{fig: trajectory_dus}
\end{figure*}

The results demonstrate that deep learning models offer superior generalization and inference performance for NLOS classification while being less prone to overfitting than classical machine learning methods. Notably, our attention-enhanced network, which integrates temporal and spatial information from all satellite observations, achieves the best performance even on out-of-distribution data. Moreover, it effectively detects NLOS in real-world vehicle localization, as shown in Fig.\,\ref{fig: trajectory_dus}.

\textbf{Take-Away Message:}
Our study on learning-based methods for GNSS outlier detection suggests their effectiveness in measurement pre-processing. However, not all methods explored can be seamlessly integrated into the state estimator, especially within the back-end optimization process.

\section{Online Learning for Noise Distribution 
Approximation}\label{sec: gmm}
\begin{figure}[!h]
    \centering
    \includegraphics[width=0.48\textwidth]{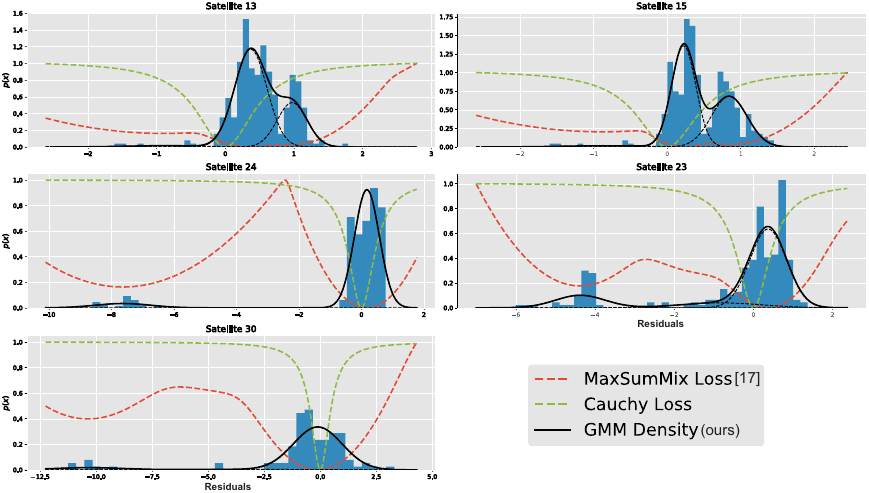}
    \caption{Demonstration of the online-approximated GMM distributions for different satellites while driving through a tree-rich area in Aachen.}
    \label{fig: gmm}
\end{figure}
We further investigate the online learning paradigm by employing variational Bayesian inference to train a Gaussian mixture model (GMM) that approximates the pseudorange noise distribution using past residual samples\footnote{This work is currently in preparation for publication.}, as illustrated in Fig.\,\ref{fig: schema}. Fig.\,\ref{fig: gmm} depicts the fitted mixture distribution alongside reference loss functions. To handle the high nonlinearity and non-convexity of the online-trained GMM, we extend the proposed FGO into a discrete-continuous optimization framework, integrating the GMM into least-square solvers. This enhancement improves both the accuracy and stability of such methods.

As shown in Table \ref{tab4:7}, our method, which incorporates distribution-shift elimination, improves accuracy by $\SI{40}{\percent}$ compared to the M-estimator and by $\SI{11}{\percent}$ the naive GMM, initially introduced in \cite{phd_Pfeifer}.

\begin{table}[!ht]
\caption{\label{tab4:7} Noise model statistics from a measurement campaign conducted along an $\SI{11}{km}$ route in Aachen.}
    \centering
    \vspace{-0.3cm}
\begin{tabular} {c|c|c|c|c}
    \hline\hline
    \multirow{2}{*}{\textbf{Noise Model}} & \multicolumn{2}{c|}{\textbf{2D Error (m)}} & \multicolumn{2}{c}{\textbf{Computation Time (ms)}}\\
    \cline{2-5}	
     & $\mathrm{mean}$ & $\mathrm{std}$ & $\mathrm{mean}$ & $\mathrm{std}$\\
    \hline
    Gaussian & $22.31$ & $28.49$ & $12.33$ & $5.16$  \\
    \hline
    M-estimator (Cauchy) & $0.90$ & $0.72$ & $\bm{11.93}$ & $4.88$  \\
    \hline
    GMM w. MPMA & $\bm{0.54}$ & $\bm{0.33}$ & $18.27$ & $6.17$  \\
    \hline
    GMM w/o. MPMA & $0.61$ & $0.36$ & $17.87$ & $\bm{4.63}$  \\
    \hline\hline
\end{tabular}
\end{table}
    
\textbf{Take-Away Message:}
Our research outcomes validate that online noise distribution approximation using past residual samples enhances both the accuracy and robustness of GNSS-based localization approaches. However, further studies on sample efficiency and augmentation, considering driving contexts such as urban and open-sky environments, are needed. This motivates us to integrate GNSS NLOS prediction with this approach in future work (see Fig.\,\ref{fig: schema}).

\section{Conclusion and Future Work}
With the outcomes from this research, we aim to explore the role of various learning-based methods in addressing inconsistent noise models in GNSS observations, enhancing robustness in challenging scenarios for robot navigation. 
Our current results motivate future work to integrate both learning paradigms in a federated system with more experimental studies, enabling their mutual enhancement and further improving overall performance, as shown in Fig.\,\ref{fig: schema}. 

\section*{Acknowledgments}
The presented research was supported by the German Federal Ministry of Economic Affairs and Climate Action (BMWK) under Projects 50NA2103 (Firefly) and 19F1150B (Museas). The work was conducted at the Institute of Automatic Control (IRT), RWTH Aachen University under the supervision of Prof. Dr.-Ing. Heike Vallery and Prof. Dr.-Ing. Dirk Abel. The author thanks Robin Taborsky from IRT for his great support in measurement campaigns. 
\bibliographystyle{IEEETran.bst}
\bibliography{reference}
\addtolength{\textheight}{-12cm}   




\end{document}